\documentclass[conference]{IEEEtran}
\IEEEoverridecommandlockouts
\usepackage{amsmath,amssymb,amsfonts}
\usepackage{algorithmic}
\usepackage[tablename=Table]{caption}
\usepackage[figurename=Figure]{caption}
\usepackage{graphicx}
\usepackage[backend=bibtex,style=ieee,natbib=true]{biblatex} 
\usepackage{graphics}
\usepackage{float}
\usepackage{dirtytalk} % quotes
\usepackage[pdftex,dvipsnames]{xcolor}
\usepackage[colorinlistoftodos,prependcaption,textsize=normal]{todonotes}
\usepackage[capitalise]{cleveref}
\usepackage{enumitem}
\usepackage{array}
\usepackage{arydshln}
\usepackage[autostyle=true]{csquotes} 
\usepackage{tabularx,ragged2e,booktabs,caption}
\usepackage{textcomp}
\usepackage{xcolor}
\renewcommand\IEEEkeywordsname{Keywords}
\makeatletter
\renewcommand{\fnum@figure}{Fig. \thefigure}
\makeatother
\begin{document}
\title{Comparison of Recurrent Neural Network Architectures for Wildfire Spread Modelling\\
}
% \def\ps@IEEEtitlepagestyle{%
%   \def\@oddfoot{\mycopyrightnotice}%
%   \def\@evenfoot{}%
% }
% \def\mycopyrightnotice{%
%   {\footnotesize\hfill The copyright belongs to me!\hfill}%   << here
%   \gdef\mycopyrightnotice{}% just in case
% }

% \IEEEpubid{\begin{minipage}{\textwidth}\ \\[12pt]
%   978-1-7281-4162-6/20/\$31.00 \copyright 2020 IEEE
% \end{minipage}} 

\newcommand\copyrighttext{%
  \footnotesize \textcopyright 2020 IEEE. Personal use of this material is permitted.  Permission from IEEE must be obtained for all other uses, in any current or future media, including reprinting/republishing this material for advertising or promotional
  purposes, creating new collective works, for resale or redistribution to servers or lists, or reuse of any copyrighted component of this work in other works.}
\newcommand\arxivcopyrightnotice{%
\begin{tikzpicture}[remember picture,overlay]
\node[anchor=south,yshift=10pt] at (current page.south) {\fbox{\parbox{\dimexpr\textwidth-\fboxsep-\fboxrule\relax}{\copyrighttext}}};
\end{tikzpicture}%
}
\def\BibTeX{{\rm B\kern-.05em{\sc i\kern-.025em b}\kern-.08em
    T\kern-.1667em\lower.7ex\hbox{E}\kern-.125emX}}

\author{\IEEEauthorblockN{Rylan Perumal}
\IEEEauthorblockA{\textit{School of Computer Science and Applied Mathematics} \\
\textit{University of the Witwatersrand}\\
Johannesburg, South Africa \\
rylan.perumal@gmail.com}
\and
\IEEEauthorblockN{Terence L van Zyl}
\IEEEauthorblockA{\textit{School of Computer Science and Applied Mathematics} \\
\textit{University of the Witwatersrand}\\
Johannesburg, South Africa\\
Terence.VanZyl@wits.ac.za}
}

% \usepackage{graphicx}
% \usepackage{floatrow}
% \usepackage{amssymb}
% \usepackage{amsmath}

% custom column size + centering

\newcolumntype{P}[1]{>{\centering\arraybackslash}m{#1}}

% \newcommandx{\note}[2][1=]{\todo[linecolor=OliveGreen,backgroundcolor=OliveGreen!25,bordercolor=OliveGreen,#1]{#2}}
% \newcommandx{\tocite}[1]{\todo[bordercolor=white,color=yellow]{#1}}
% \newcommand{\tochange}[1]{\colorbox{Apricot}{#1}}
\newcommand{\code}{\texttt}

\maketitle

\arxivcopyrightnotice

\IEEEpubidadjcol

\begin{abstract}
Wildfire modelling is an attempt to reproduce fire behaviour. Through active fire analysis, it is possible to reproduce a dynamical process, such as wildfires, with limited duration time series data. Recurrent neural networks (RNNs) can model dynamic temporal behaviour due to their ability to remember their internal input. In this paper, we compare the Gated Recurrent Unit (GRU) and the Long Short-Term Memory (LSTM) network. We try to determine whether a wildfire continues to burn and given that it does, we aim to predict which one of the 8 cardinal directions the wildfire will spread in. Overall the GRU performs better for longer time series than the LSTM. We have shown that although we are reasonable at predicting the direction in which the wildfire will spread, we are not able to asses if the wildfire continues to burn due to the lack of auxiliary data.
\\
\end{abstract}

\begin{IEEEkeywords}
machine learning, wildfire spread modelling, recurrent neural networks
\end{IEEEkeywords}
\section{INTRODUCTION}
A recurrent neural network (RNN) is a class of artificial neural networks which contain feedback connections that allow information to persist in the network. This allows them to model temporal dynamic behaviour for a time series. Unlike feedforward neural networks, RNNs remember their inputs due to their internal memory \cite{RNNVarious}.
\\
It has been observed by \citet{Bengio} that it is difficult to train RNNs to capture long-term dependencies, due to the vanishing gradient problem. More sophisticated RNN architectures that will be investigated include the Long Short-Term Memory (LSTM) and Gated Recurrent Unit (GRU), these architectures are designed such that they overcome the vanishing gradient problem \cite{GRU, RC, SpaceOdyssey, cho}. 
\\
\\
Wildfire spread modelling is an attempt to reproduce fire behaviour. Key factors which contribute to this include the amount of combustible material around it, the weather conditions and the topography of the Earth's surface. It is often the case when trying to model dynamical processes, such as wildfires, that only limited duration time series data is available and the initial model which generated the data is unavailable. Previously, forest fire modelling has been attempted using a software system which simulates the behaviour of a fire. The results are then compared to real-world data obtained through active fire analysis\footnote{Active fire analysis consists of using remote sensing through a spectral signal which is observable from space. This spectral signal is picked up, by a satellite, from the radiation given off by the fire.} \cite{Kozik2013, Kozik2014, Chaos2017}.
\\
\\
In this paper, we investigate the performance of the LSTM and GRU architectures with Logistic Regression (LR) as our baseline. Fire data is obtained through active fire analysis. We then use this data to construct wildfires, the learning material, by finding the K-Nearest Neighbours of each point. We then graph each point and its neighbours and then extract sub-graphs which represent wildfires. The RNN's and the baseline are evaluated based on their averaged accuracy, precision and recall. We evaluate two different cases for wildfire spread modelling which include binary classification, where we try to predict whether or not the wildfire continues to burn and multiclass classification, where we try to predict in which one of the $8$ cardinal directions the wildfire will move in given that it continues to burn.
\\
\\
The rest of this paper follows the following structure. Section \ref{background} covers the relevant background for wildfire spread modelling and RNNs. Section \ref{methodology} goes over our methodology to construct wildfires, the experiment setup where we discuss how we generated input and output for the binary and multiclass cases and the models that we used for training and classification. Section \ref{results} shows the results we obtained from each model for each type of classification as well as a discussion. Finally, Section \ref{conclusion} concludes what we have implemented, the results we have produced and future improvements.

% The focus of this research is to compare different recurrent neural network architectures on the task of wildfire spread modelling using the aforementioned approaches. In particular, using the reservoir computing technique, the ESN, and evaluating the performance of this architecture in comparison to the LSTM and GRU architectures.

\section{Background} \label{background}

\subsection{Wildfire Spread Modelling}
Modelling dynamic processes on the Earth's surface is characterised by a high degree of a priori uncertainty because many parameters of the Earth's surface that affect the process' evolution are either known with certain error from the very beginning or change with time. Such processes may include fires, floods, oil spillage etc \cite{Kozik2013}.

Mathematical models of forest fires are usually divided into three groups: empirical (statistical); semi-empirical (semi-physical) and physical (analytical) models. Kozik, Nezhevenko and Feoktistov \cite{Kozik2014} use a basic mathematical model, namely, Rothermel's semi-empirical model, for creeping fire evolution, which was well approved in most forest services in North America. Rothermel's model is based on simple models and determines the velocity of forest fire propagation by taking into account the following information as input: the amount of combustible material; moisture content; effective heat of ignition, $Q_{eff}$; the heat flux, $I_{R}$, released from the burning material onward; the wind coefficient, $\phi_{wind}$ and the slope coefficient, $\phi_{slope}$. This model can characterise the process of heat transfer of the fire, give feedback on the heat release intensity, burning time and the velocity of the fire propagation \cite{rothermel, Kozik2014, Kozik2013}. 
\\
\\
Kozik, Nezhevenko and Feoktistov \cite{Kozik2014} simulate forest fire behaviour using a non-completely connected RNN and input based on Rothermel's model. It is shown by \cite{Kozik2013} that modelling of a forest fire ensures fire data assimilation through comparing the simulated fire spread to the real observations obtained through active fire analysis. Data assimilation offers the possibility to reduce a priori uncertainty, which is always inherent when modelling dynamic processes \cite{Kozik2013, Kozik2014}.

\subsection{Recurrent Neural Networks}

A recurrent neural network (RNN) is a neural network with feedback connections. These connections allow information to persist in the network. More formally, RNNs can handle a sequence of inputs which are of arbitrary length. RNNs handle these inputs by having a recurrent hidden state whose activation at each time is dependent on that of the previous time. A common problem with normal RNNs is that they suffer from gradient-based training. Thus other approaches have been made to reduce the negative impacts of this problem which include the GRU and LSTM architectures \cite{RNNVarious, Hochreiter, cho, GRU, SpaceOdyssey}.
\\
\subsubsection{Long Short-Term Memory}
The Long Short-Term Memory (LSTM) architecture is effective at capturing long-term temporal dependencies. The main idea behind the LSTM architecture is a memory cell, which can maintain its state over time and gating units, which regulate the information flow into and out of the cell. The LSTM architecture used in the literature, known as the vanilla LSTM, was originally described by \cite{LSTM_graves}. It features: an input, output and forget gates; block input; a single cell (the constant error carousel (CEC)), which provides short term memory storage by recirculating activations (sigmoids) and error signals; input and output activation functions and peephole connections. The forget gate enables the LSTM to reset its state, peephole connections allow communication between the CEC and the different gates, which in turn allow for better learning and finally the last addition, gradient-based learning. This overall allowed gradients to be checked using finite difference methods, thus making practical implementations more reliable  \cite{SpaceOdyssey, LSTMGers, LSTM_graves}.
\\
\subsubsection{Gated Recurrent Unit}
A Gated Recurrent Unit (GRU) is a variant of the LSTM. The GRU can adaptively capture dependencies of different time scales. Like the LSTM, the GRU has gating units that allow information to flow in and out, however, there is no separate memory cell. Unlike the LSTM, the GRU combines the forget and input gates into a single update gate which controls how much memory the hidden unit can hold. The GRU also contains a reset gate which resets the memory of the cell. An empirical comparison conducted by \cite{GRU} between the LSTM and GRU networks (which have been configured with the same hyper-parameters) shows that on some datasets the GRU performs better in overall generalisation. Jozefowicz, Zaremba and Sutskever \cite{Zaremba} show that the GRU outperformed the LSTM at predicting the next character in a synthetic XML dataset and the Penn TreeBank dataset \cite{RNNVarious, cho, RNN_survey, E_rnn, GRU}.

\section{Methodology} \label{methodology}
\subsection{Wildfire Generation}

\subsubsection{Data Preparation}
An active fire dataset from the Visible Infrared Imaging Radiometer Suite (VIIRS) for the years 2012--2014 was used. The data points within a bounding box of South Africa were extracted. Each data point contained the following variables: latitude; longitude; time; date and fire radiative power (FRP). To this extracted dataset the elevation was added using a GPS Visualiser tool\footnote{https://www.gpsvisualizer.com/elevation}. The time and date variables were converted to one-hot encoded vectors which represent an hour of the day and the week of the year respectively. All of the continuous variables in the dataset were normalised. The dimension of a given point is $\in \mathbb{R}^{80}$ which includes the one-hot encoded vectors, latitude, longitude, FRP and elevation.
\\
\subsubsection{K-Nearest Neighbours}
\begin{figure}[t]
\begin{center}
    \includegraphics[scale=0.3]{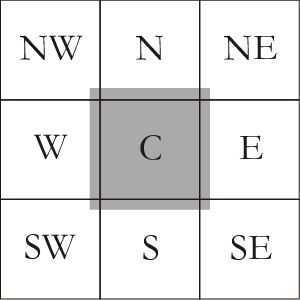}
    \caption{The $8$ cardinal directions surrounding a given point $C$ \cite{cardDir}.}
    \label{figure1}
\end{center}
\end{figure}

To construct wildfires, we grouped points that were close together in space and in time. For a given point $C$, we wanted to know which points (nearest neighbours) were positioned at the $8$ cardinal directions as seen in Fig. \ref{figure1}. We used K-Nearest Neighbours from Scikit-Learn\footnote{https://scikit-learn.org/stable/modules/neighbors.html}, where $K=8$. As a precaution of including neighbours that are not valid in time and space, for each point, $C$, the corresponding set of neighbours, $C_{neighbours}$, were filtered to only consider neighbours within a spatial radius, $s_r=375m$ (meters) as the VIIRS dataset only offers $375m$ of spatial resolution and simultaneously only considering points that were within a temporal radius, $t_r=6h$ (hours). This is based on the assumption that any point in $C_{neighbours}$ where $t_r > 6$ with respect to $C$ is not a neighbour to $C$. This newly constructed dataset contains each point as well as its filtered $K$ nearest neighbours, where $K$ is now $ \leq 8$.
\\
\subsubsection{Wildfire Graphing}
The dataset obtained from KNN was graphed using NetworkX\footnote{https://networkx.github.io/}. Connections were made between each point and its neighbours. This formed a massive graph network with many disjoint sub-components, each of which assimilates a wildfire. These components were then extracted and formed the final wildfire dataset. Table \ref{table1} shows us the statistics of the wildfire dataset which includes the total number of wildfires, the mean wildfire length, the standard deviation of wildfire length and the length of the shortest and longest wildfire.

\begin{table}[b]
\begin{minipage}{.5\textwidth}
\centering
\renewcommand{\arraystretch}{2}
\caption{Table showing the statistics of the constructed wildfire dataset.}
\begin{tabular}[t]{|c|c|} 
\hline
\multicolumn{2}{|c|}{Wildfire Dataset Statistics} \\
\hline
Number of Wildfires & 272894 \\ 
\hline
Mean of Wildfire Length & 1.905494 \\
\hline
Standard Deviation of Wildfire Length & 3.248123 \\
\hline
Shortest Wildfire Length & 1 \\
\hline
Longest Wildfire Length & 334 \\
\hline
\end{tabular}
\label{table1}
\end{minipage}%
\end{table}

% \begin{figure}[H]
% \begin{center}
%     \includegraphics[scale=0.4]{Figures/wildfire_distribution.png}
%     \caption{Wildfire distribution for length of a wildfire.}
%     \label{figure_w}
% \end{center}
% \end{figure}

\subsection{Experiment Setup}
\subsubsection{Input Generation}
We consider wildfires of length $\bar{l} = \{2, 3, 4, 5, 6, 7, 8\}$ for the rest of this article.
\\
\paragraph{Binary}
In the binary case, we want to be able to predict whether a wildfire continues to burn. Using wildfires of length $l_w$, where $l_{w} \in \bar{l}$ and $l_{w}-1$, we remove the last point in the fires of length $l_w$, where we now have two sets of input with the same length. Wildfires originally of length $l_w$ correspond to label $1$ (burning) and wildfires of length $l_{w} -1$ correspond to label $0$ (not burning). 
\paragraph{Multiclass}
In the multiclass case we want to be able to predict which direction the next point in a wildfire will move provided that it continues to burn. The cardinal direction $d_{i+1}$, where $d \in [1, 2, \dots, 8]$ corresponding to Fig. \ref{figure1} of the point $C_{i+1}$, is calculated using points $C_{i}$ and $C_{i+1}$ where $i \in [0, \dots, (l_{w}-1)]$. Each $d_{i+1}$ is appended to the input vector up until $i=(l_{w}-2)$. For the first point in the wildfire, $C_{0}$, the value of $d_{0} = 0$, as this point has no previous direction. Thus making the dimension of the input, $\Bar{x} \in \mathbb{R}^{ (80\times(l_w-1))+(l_w-1)}$. When $i=(l_{w}-1)$ the value of $d_{i+1}$ is the corresponding prediction label.
\\
\subsubsection{Training and Classification}
Training the models consisted of using a train test split for wildfires of length $l_w$, where the test size was $30\%$ of the total number of wildfires of length $l_w$. The training set went through a $10$-fold cross validation where the best model was selected from the validation set. The test set was then evaluated using the best model where the accuracy, precision and recall were recorded. These results were further averaged $10$ times for each $l_w \in \Bar{l}$.  
\\
\paragraph{Logistic Regression (LR)}
This model was used as our baseline. The main purpose for this was to ensure that we are doing better than a basic approach. Rmsprop\footnote{The idea behind rmsprop was to keep a moving average of the squared gradients for each parameter. The gradient is divided by the root mean square of the gradient \cite{rmsprop}.} is used to optimise the parameters with a mean square error (MSE) loss function and trained for $300$ epochs.

\paragraph{Long Short-Term Memory (LSTM) and Gated Recurrent Unit (GRU)}
These networks were implemented using the python library Keras\footnote{https://keras.io/}. The input was fed into a dense layer matching the dimensions of input vector $\Bar{x}$ with linear activation, followed by two hidden layers consisting of $128$ and $256$ LSTM/GRU units (neurons) respectively, both with rectified linear (ReLU) activation. A dropout layer was added to prevent the model from overfitting the training data. Finally, an output layer was added with sigmoid activation, where the number of units is equal to the number of classes being predicted. The parameters of this model were optimised using rmsprop with a MSE loss function and each model was trained for $20$ epochs.

\section{Results} \label{results}
In this section we discuss the results from our models which are compared based on the accuracy, recall, and precision which are briefly explained below:  
\begin{itemize}
    \item \textbf{Accuracy: } Average number of correct predictions.
    \item \textbf{Precision:} Out of the classes that were predicted as true, how many of them actually true.
    \item \textbf{Recall: } Out of the total number of true classes how many were actually predicted as true.
\end{itemize}

\subsection{Binary}
\begin{figure}
\begin{center}
    \includegraphics[scale=0.45]{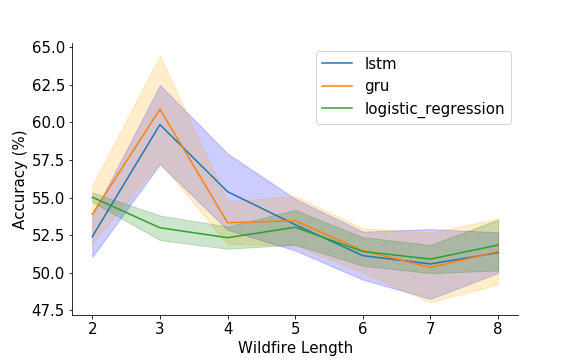}
    \caption{Accuracy plot for binary classification for different wildfire lengths.}
    \label{figure3}
\end{center}
\end{figure}

\begin{figure}[b]
\begin{center}
    \includegraphics[scale=0.45]{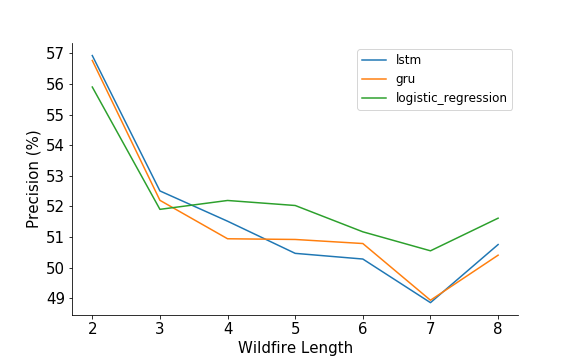}
    \caption{Precision plot for binary classification for different wildfire lengths.}
    \label{figure6}
\end{center}
\end{figure}

\begin{figure}
\begin{center}
    \includegraphics[scale=0.45]{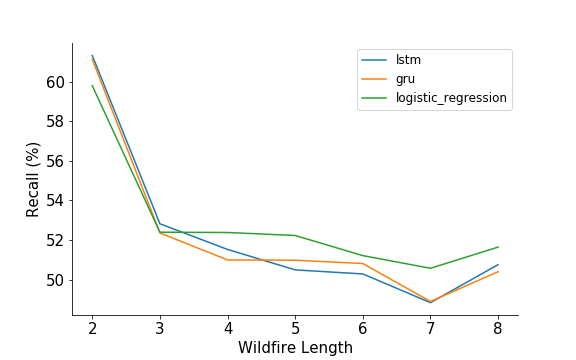}
    \caption{Recall plot for binary classification for different wildfire lengths.}
    \label{figure7}
\end{center}
\end{figure}

Fig. \ref{figure3} and Table \ref{tab:my-table2} shows us the accuracy between each model for wildfires of different length, where the shaded regions illustrate the accuracy one standard deviation from the mean. When $l_w = 2$ the LR model does better than the RNNs. This is expected as we have a short time series. The RNNs do better than the baseline for $l_{w} = \{3, 4, 5\}$. However, the RNNs performance decreases quite sharply for $l_w \geq 3$. For $l_w > 5$, the models' performance in comparison to each other is relatively close. From the accuracy results we can see that all $3$ models are not good at determining whether a wildfire continues to burn for $l_{w} > 4$. Further, it is not clear as to which RNN architecture is better in terms of accuracy alone.
\\
\\
Fig. \ref{figure6} and Fig. \ref{figure7} show the precision and recall plots respectively for each model for binary classification. Table \ref{tab:my-table2} shows the values of precision and recall corresponding to Fig. \ref{figure6} and Fig. \ref{figure7}. When $l_w = 2$, the RNNs are more precise and have better recall. However, when $l_w \geq 3$ the baseline's performance is similar to the performance of the RNN architectures. This tells us that RNNs, possibly, cannot handle long time series binary classification well. Further, the task of predicting whether a fire continues to burn, for wildfires of length $l_{w} > 4$, is quite difficult. It is not clear as to which RNN has better precision or recall as the graphs in Fig. \ref{figure6} and Fig. \ref{figure7} overlap each other at different wildfire lengths.

\begin{table}[h]
\caption{Binary classification results for models implemented.}
\label{tab:my-table2}
\resizebox{\columnwidth}{!}{
\renewcommand{\arraystretch}{2}
\begin{tabular}{|l|l|l|l|l|l|l|l|l|l|l|l|}
\hline
\multicolumn{1}{|c|}{} & \multicolumn{5}{c|}{Accuracy}                                                        & \multicolumn{3}{c|}{Precision}                                                 & \multicolumn{3}{c|}{Recall}                                                    \\ \hline
$l_{w}$                & \multicolumn{2}{c|}{\textbf{LR}}     & \multicolumn{2}{c|}{\textbf{LSTM}}   & \multicolumn{1}{c|}{\textbf{GRU}} & \multicolumn{1}{c|}{\textbf{LR}} & \multicolumn{1}{c|}{\textbf{LSTM}} & \multicolumn{1}{c|}{\textbf{GRU}} & \multicolumn{1}{c|}{\textbf{LR}} & \multicolumn{1}{c|}{\textbf{LSTM}} & \multicolumn{1}{c|}{\textbf{GRU}} \\ \hline
2                      & \multicolumn{2}{l|}{0.5502} & \multicolumn{2}{l|}{0.5241} & 0.5389                   & 0.5590                  & 0.5692                    & 0.5676                   & 0.5980                  & 0.6132                    & 0.6113                   \\ \hline
3                      & \multicolumn{2}{l|}{0.5299} & \multicolumn{2}{l|}{0.5985} & 0.6088                   & 0.5190                  & 0.5250                    & 0.5219                   & 0.5239                  & 0.5282                    & 0.5236                   \\ \hline
4                      & \multicolumn{2}{l|}{0.5233} & \multicolumn{2}{l|}{0.5539} & 0.5332                   & 0.5219                  & 0.5151                    & 0.5093                   & 0.5238                  & 0.5152                    & 0.5099                   \\ \hline
5                      & \multicolumn{2}{l|}{0.5302} & \multicolumn{2}{l|}{0.5319} & 0.5346                   & 0.5202                  & 0.5046                    & 0.5091                   & 0.5223                  & 0.5049                    & 0.5098                   \\ \hline
6                      & \multicolumn{2}{l|}{0.5141} & \multicolumn{2}{l|}{0.5113} & 0.5147                   & 0.5116                  & 0.5027                    & 0.5078                   & 0.5121                  & 0.5028                    & 0.5081                   \\ \hline
7                      & \multicolumn{2}{l|}{0.5091} & \multicolumn{2}{l|}{0.5058} & 0.5035                   & 0.5054                  & 0.4885                    & 0.4892                   & 0.5057                  & 0.4883                    & 0.4889                   \\ \hline
8                      & \multicolumn{2}{l|}{0.5184} & \multicolumn{2}{l|}{0.5134} & 0.5141                   & 0.5161                  & 0.5075                    & 0.5040                   & 0.5164                  & 0.5075                    & 0.5040                   \\ \hline
\end{tabular}
}
\end{table}

\subsection{Multiclass}
\begin{figure}[t]
\begin{center}
    \includegraphics[scale=0.45]{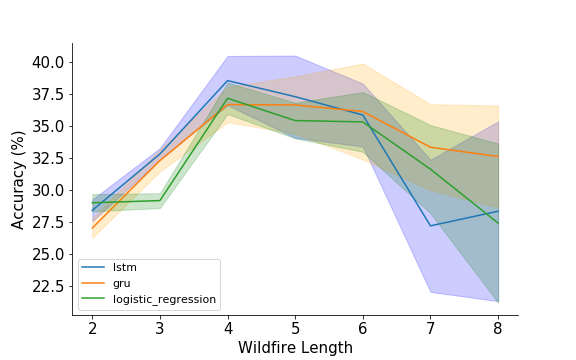}
    \caption{Accuracy plot for multiclass classification for different wildfire lengths.}
    \label{figure2}
\end{center}
\end{figure}

\begin{figure}
\begin{center}
    \includegraphics[scale=0.45]{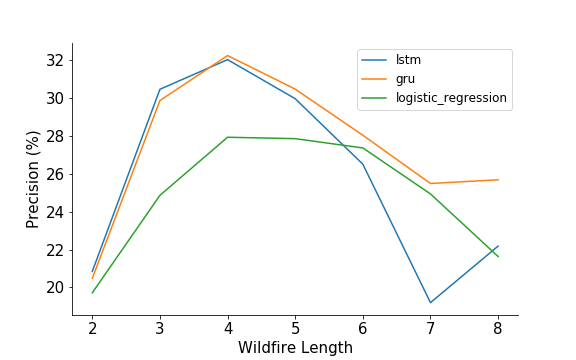}
    \caption{Precision plot for multiclass classification for different wildfire lengths.}
    \label{figure4}
\end{center}
\end{figure}

\begin{figure}
\begin{center}
    \includegraphics[scale=0.45]{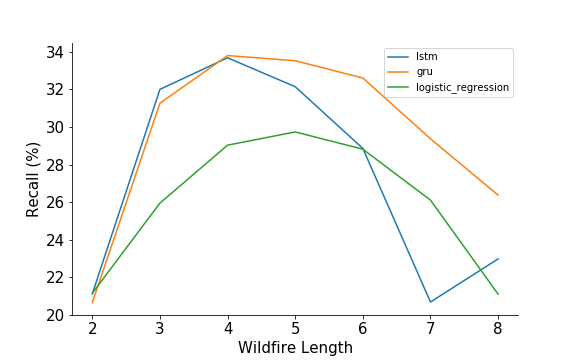}
    \caption{Recall plot for multiclass classification for different wildfire lengths.}
    \label{figure5}
\end{center}
\end{figure}

Fig. \ref{figure2} and Table \ref{tab:my-table} shows us the accuracy between each model for wildfires of different lengths and the shaded regions illustrate the accuracy one standard deviation from the mean. It is clear to see that when $l_{w}=2$ the LR model does better than the GRU and LSTM. This is because there is a single time series input which results in the RNNs not being able to fit any additional autocorrelation. As $l_{w}$ increases, the RNNs out-perform LR due to longer time series input. The LSTM network is slightly better than the GRU (in terms of accuracy) up until $l_w \geq 6$, from this point onward the GRU performs the best. It is noteworthy that, the LSTM has the highest accuracy for $l_w = 4$. For longer time series the GRU is also more stable than the LSTM which can be seen by the large area of the blue (LSTM) shadow in comparison to the orange (GRU) shadow.
\\
\\
Fig. \ref{figure4} and Fig. \ref{figure5} show the precision and recall respectively of each model for multiclass classification. Table \ref{tab:my-table} shows the values of precision and recall corresponding to Fig. \ref{figure4} and Fig. \ref{figure5}. The precision of the RNNs outperforms the baseline. The LSTM's precision is better by a small margin in comparison to the GRU until $l_w \geq 4$, the precision of the GRU surpasses the LSTM by a significant margin. Further, we can observe that for $l_w \geq 6$ LR is more precise than the LSTM. From Fig. \ref{figure4} we can infer that the GRU is most precise in predicting the correct direction of wildfire spread. We find a similar trend in Fig. \ref{figure5}, the LSTM has better recall for short time series compared to the GRU and LR however, as the length of the time series grows the baseline has a much better recall in comparison to the LSTM. The GRU and LR surpasses the LSTM in recall for $l_w \geq 4$ and for $l_w \geq 6$ respectively. This tells us that the GRU has a better recall in comparison to LR and more importantly the LSTM. The multiclass classification results coincides with the literature which states that the GRU does better than the LSTM for longer time series with certain datasets \cite{cho, E_rnn, RNN_survey}.

\begin{table}[h]
\caption{Multiclass classification results for models implemented.}
\label{tab:my-table}
\resizebox{\columnwidth}{!}{
\renewcommand{\arraystretch}{2}
\begin{tabular}{|l|l|l|l|l|l|l|l|l|l|l|l|}
\hline
\multicolumn{1}{|c|}{} & \multicolumn{5}{c|}{Accuracy}                                                        & \multicolumn{3}{c|}{Precision}                                                 & \multicolumn{3}{c|}{Recall}                                                    \\ \hline
$l_{w}$                & \multicolumn{2}{c|}{\textbf{LR}}     & \multicolumn{2}{c|}{\textbf{LSTM}}   & \multicolumn{1}{c|}{\textbf{GRU}} & \multicolumn{1}{c|}{\textbf{LR}} & \multicolumn{1}{c|}{\textbf{LSTM}} & \multicolumn{1}{c|}{\textbf{GRU}} & \multicolumn{1}{c|}{\textbf{LR}} & \multicolumn{1}{c|}{\textbf{LSTM}} & \multicolumn{1}{c|}{\textbf{GRU}} \\ \hline
2                      & \multicolumn{2}{l|}{0.2898} & \multicolumn{2}{l|}{0.2839} & 0.2700                   & 0.1972                  & 0.2084                    & 0.2048                   & 0.2115                  & 0.2111                    & 0.2065                   \\ \hline
3                      & \multicolumn{2}{l|}{0.2915} & \multicolumn{2}{l|}{0.3280} & 0.3229                   & 0.2485                  & 0.3043                    & 0.2984                   & 0.2594                  & 0.3199                    & 0.3125                   \\ \hline
4                      & \multicolumn{2}{l|}{0.3716} & \multicolumn{2}{l|}{0.3855} & 0.3667                   & 0.2791                  & 0.3200                    & 0.3221                   & 0.2902                  & 0.3367                    & 0.3379                   \\ \hline
5                      & \multicolumn{2}{l|}{0.3542} & \multicolumn{2}{l|}{0.3729} & 0.3664                   & 0.2783                  & 0.2994                    & 0.3044                   & 0.2973                  & 0.3213                    & 0.3351                   \\ \hline
6                      & \multicolumn{2}{l|}{0.3531} & \multicolumn{2}{l|}{0.3585} & 0.3612                   & 0.2735                  & 0.2649                    & 0.2801                   & 0.2881                  & 0.2885                    & 0.3259                   \\ \hline
7                      & \multicolumn{2}{l|}{0.3161} & \multicolumn{2}{l|}{0.2717} & 0.3333                   & 0.2492                  & 0.1920                    & 0.2547                   & 0.2610                  & 0.2068                    & 0.2936                   \\ \hline
8                      & \multicolumn{2}{l|}{0.2739} & \multicolumn{2}{l|}{0.2832} & 0.3260                   & 0.2163                  & 0.2218                    & 0.2567                   & 0.2110                  & 0.2297                    & 0.2636                   \\ \hline
\end{tabular}
}
\end{table}

% \\
% \\
% The above analysis shows us that RNNs are better at multiclass classification than binary. Further that the GRU outperforms the LSTM when it comes to this application. However for binary classification the RNNs in general do not perform well. This could be due to the fact that we are using the exact same models to do binary classification. 

\section{Conclusion} \label{conclusion}
We have compared two different approaches for wildfire spread modelling, that is, trying to predict whether the wildfire continues to burn (binary classification) and which one of the $8$ cardinal directions, shown in Fig. \ref{figure1}, will the wildfire move provided that it continues to burn (multiclass classification). For the case of multiclass classification, the GRU is more stable, precise and has better recall than the LSTM for longer time series. Therefore the GRU is a better RNN architecture for modelling the direction in which a wildfire spreads given that it continues to burn. Although we are reasonable at predicting the direction of wildfire spread, we are not able to asses if the next point will burn. This is due to the lack of auxiliary data which includes features such as moisture content, wind direction, wind speed etc. Future work that can be done to improve results includes: implementing a different model which is optimised for binary classification, evaluating the models at different epochs and adding extra information to the dataset. Important information that could be added includes the NDVI (Normalised Difference Vegetation Index), Earth data from MODIS (Moderate Resolution Imaging Spectroradiometer) and possibly an elevation gradient at a particular point on the Earth's surface. Adding these attributes to the dataset and implementing a more optimised model for the binary problem can have a significantly positive effect on overall accuracy, precision and recall for wildfire spread modelling.

% hU network is builthe GRere we have to describe how the binary and multiclass data were generated, then we have to describe the LSTM, GRU and Logistic Regression Classifiers
% ACKNOWLEDGMENT 
\section*{acknowledgment}
I would like to thank Prof. van Zyl for all of his intellectual insight and motivation throughout. It was a great pleasure being supervised by him.

\printbibliography[heading=bibintoc]

@article{RNNVarious,
  title = {Recurrent Neural Network and its Various Architecture Types},
  author = {Trupti Katte},
  year = {2018},
  journal = {International Journal of Research and Scientific Innovation},
  volume = {5},
  issue = {3},
}

@article{Kozik2014,
 author = {Kozik, V.I. and Nezhevenko, E.S. and Feoktistov, A.S.},
 title = {Studying the method of adaptive prediction of forest fire evolution on the basis of recurrent neural networks},
 journal = {Optoelectronics, Instrumentation and Data Processing},
 year = {2014},
 day = {01},
 pages = {395--401},
}

@inproceedings{Zaremba,
  title={An empirical exploration of recurrent network architectures},
  author={Jozefowicz, Rafal and Zaremba, Wojciech and Sutskever, Ilya},
  booktitle={International Conference on Machine Learning},
  pages={2342--2350},
  year={2015}
}

@article{Kozik2013,
author = {Kozik, V. I.
and Nezhevenko, E. S.
and Feoktistov, A. S.},
title = {Adaptive prediction of forest fire behavior on the basis of recurrent neural networks},
journal = {Optoelectronics, Instrumentation and Data Processing},
year = "2013",
day="01",
volume="49",
number="3",
pages="250--259",
}

@article{rothermel,
  title={A mathematical model for predicting fire spread in wildland fuels},
  author={Rothermel, Richard C.},
  journal={Res. Pap. INT-115. Ogden, UT: U.S. Department of Agriculture, Intermountain Forest and Range Experiment Station.},
  year={1972},
}

@article{LSTMGers,
 author = {Gers, Felix A. and Schraudolph, Nicol N. and Schmidhuber, J\"{u}rgen},
 title = {Learning Precise Timing with {LSTM} Recurrent Networks},
 journal = {J. Mach. Learn. Res.},
 issue_date = {3/1/2003},
 volume = {3},
 year = {2003},
 pages = {115--143},
 numpages = {29},
 publisher = {JMLR.org},
}

@article{GRU,
  title={Empirical Evaluation of Gated Recurrent Neural Networks on Sequence Modeling},
  author={Junyoung Chung and Çaglar G{\"u}lçehre and Kyunghyun Cho and Yoshua Bengio},
  journal={CoRR},
  year = {2014},
  month = {12}, 
}

@article{Bengio,
 author = {Bengio, Y. and Simard, P. and Frasconi, P.},
 title = {Learning Long-term Dependencies with Gradient Descent is Difficult},
 journal = {Trans. Neur. Netw.},
 volume = {5},
 number = {2},
 year = {1994},
 pages = {157--166},
 numpages = {10},
 publisher = {IEEE Press},
}

@inproceedings{cho,
    title = "On the Properties of Neural Machine Translation: Encoder{--}Decoder Approaches",
    author = "Cho, Kyunghyun  and
      van Merrienboer, Bart  and
      Bahdanau, Dzmitry  and
      Bengio, Yoshua",
    booktitle = "Proceedings of {SSST}-8, Eighth Workshop on Syntax, Semantics and Structure in Statistical Translation",
    year = "2014",
    address = "Doha, Qatar",
    publisher = "Association for Computational Linguistics",

    pages = "103--111",
}

@article{Hochreiter,
author = {Hochreiter, Sepp and Schmidhuber, Jürgen},
title = {Long Short-Term Memory},
journal = {Neural Computation},
volume = {9},
number = {8},
pages = {1735-1780},
year = {1997},
}

@article{SpaceOdyssey, 
author={K. {Greff} and R. K. {Srivastava} and J. {Koutník} and B. R. {Steunebrink} and J. {Schmidhuber}}, 
journal={IEEE Transactions on Neural Networks and Learning Systems}, 
title={{LSTM}: A Search Space Odyssey}, 
year={2017}, 
volume={28}, 
number={10}, 
pages={2222--2232}
}

@article{LSTM_graves,
title = "Framewise phoneme classification with bidirectional {LSTM} and other neural network architectures",
journal = "Neural Networks",
volume = "18",
number = "5",
pages = "602--610",
year = "2005",
author = "Alex Graves and Jürgen Schmidhuber",
}

@book{RNN_survey,
   title={Recurrent neural networks for short-term load forecasting: an overview and comparative analysis},
   journal={SpringerBriefs in Computer Science},
   publisher={Springer International Publishing},
   author={Bianchi, Filippo Maria and Maiorino, Enrico and Kampffmeyer, Michael C. and Rizzi, Antonello and Jenssen, Robert},
   year={2017}
}

@inproceedings{cardDir,
author = {Pfoser, Dieter and Hadzilacos, Thanasis and Faradouris, Nikos and Kyrimis, Kriton},
year = {2007},
month = {01},
pages = {287--302},
title = {Spatial Data Management Aspects in Archaeological Excavation Documentation 1},
}

@InProceedings{E_rnn,
  title = 	 {An Empirical Exploration of Recurrent Network Architectures},
  author = 	 {Rafal Jozefowicz and Wojciech Zaremba and Ilya Sutskever},
  booktitle = 	 {Proceedings of the 32nd International Conference on Machine Learning},
  pages = 	 {2342--2350},
  year = 	 {2015},
  editor = 	 {Francis Bach and David Blei},
  volume = 	 {37},
  series = 	 {Proceedings of Machine Learning Research},
  address = 	 {Lille, France},
  month = 	 {07--09 Jul},
  publisher = 	 {PMLR},
}

@article{Chaos2017,
    author = {{Pathak}, Jaideep and {Lu}, Zhixin and {Hunt}, Brian R. and
    {Girvan}, Michelle and {Ott}, Edward},
    title = "{Using machine learning to replicate chaotic attractors and calculate Lyapunov exponents from data}",
    journal = {Chaos: An Interdisciplinary Journal of Nonlinear Science},
    year = {2017},
    volume = {27},
    pages={121102}
}

@article{RC,
title = "Reservoir computing and extreme learning machines for non-linear time-series data analysis",
journal = "Neural Networks",
volume = "38",
pages = "76--89",
year = "2013",
author = "J.B. Butcher and D. Verstraeten and B. Schrauwen and C.R. Day and P.W. Haycock",
}

@misc{rmsprop,
  title={{Lecture 6.5---RmsProp: Divide the gradient by a running average of its recent magnitude}},
  author={Tieleman, T. and Hinton, G.},
  howpublished={COURSERA: Neural Networks for Machine Learning},
  year={2012}
}
% \bibliography{mybib}

\end{document}